\documentclass[10pt, conference, compsocconf]{IEEEtran}
\ifCLASSINFOpdf
\else
\fi
\usepackage{cite}
\usepackage{amsmath,amssymb,amsfonts}
\usepackage{algorithmic}
\usepackage{graphicx}
\usepackage{textcomp}
\usepackage{xcolor}
\usepackage{subfigure}
\usepackage{makecell}
\usepackage{multirow}
\def\BibTeX{{\rm B\kern-.05em{\sc i\kern-.025em b}\kern-.08em
    T\kern-.1667em\lower.7ex\hbox{E}\kern-.125emX}}

\def\BibTeX{{\rm B\kern-.05em{\sc i\kern-.025em b}\kern-.08em
    T\kern-.1667em\lower.7ex\hbox{E}\kern-.125emX}}
\usepackage{hyperref}    
\hypersetup{colorlinks=true,
linkcolor=green,
hypertexnames=false,
pdfhighlight=/N
}    

\begin{document}
%
\title{Driver Identification Based on Vehicle Telematics Data using LSTM-Recurrent Neural Network \\
\thanks{This work is support by NASA Langley Research Center and Air Force Research Laboratory and Office of the Secretary of Defence (OSD)}
}


\author{\IEEEauthorblockN{Abenezer Girma, \textit{Student Member ,IEEE}, 
 Xuyang Yan and 
Abdollah Homaifar
}
\IEEEauthorblockA{Autonomous Control and Information Technology (ACIT) Institute\\
Department of Electrical and Computer Engineering,
North Carolina A$\&$T State University\\
Email: aggirma@aggies.ncat.edu,
xyan@aggies.ncat.edu,
homaifar@ncat.edu}}

\maketitle

\begin{abstract}
Despite advancements in vehicle security systems, over the last decade, auto-theft rates have increased, and cyber-security attacks on internet-connected and autonomous vehicles are becoming a new threat. In this paper, a deep learning model is proposed, which can identify drivers from their driving behaviors based on vehicle telematics data. The proposed Long-Short-Term-Memory (LSTM) model predicts the identity of the driver based on the individual's unique driving patterns learned from the vehicle telematics data. Given the telematics is time-series data, the problem is formulated as a time series prediction task to exploit the embedded sequential information. The performance of the proposed approach is evaluated on three naturalistic driving datasets, which gives high accuracy prediction results. The robustness of the model on noisy and anomalous data that is usually caused by sensor defects or environmental factors is also investigated. Results show that the proposed model prediction accuracy remains satisfactory and outperforms the other approaches despite the extent of anomalies and noise-induced in the data.
\end{abstract}

\section{Introduction}
Although the technological improvement of automobile technologies are advancing, the vehicles, security problem is not sufficiently addressed. For the last decade, auto theft rates have increased around the globe. According to an FBI Crime Report, in 2017 there has been an estimated 773,139 motor vehicle thefts in the USA, a 10.4$\%$ increase when compared with the 2013 report \footnote{https://ucr.fbi.gov/crime-in-the-u.s/2017/crime-in-the-u.s.-2017/topic-pages/motor-vehicle-theft}. Secondly, connected and autonomous cars are linked to the internet, which increases their vulnerability for cyber-attacks more than ever. In 2015, Jeep recalled 1.4 million connected vehicles after hackers remotely hacked and controlled the 2014 model Jeep car over the Internet\footnote{https://www.wired.com/2015/07/jeep-hack-chrysler-recalls-1-4m-vehicles-bug-fix/}. Thirdly, in shared mobility and insurance companies, identifying the car operator is vital in preventing dangers caused by unauthorized drivers. 

\begin{figure}[!h]
	\centering
	\includegraphics[width=3.6in,height=2.6in]{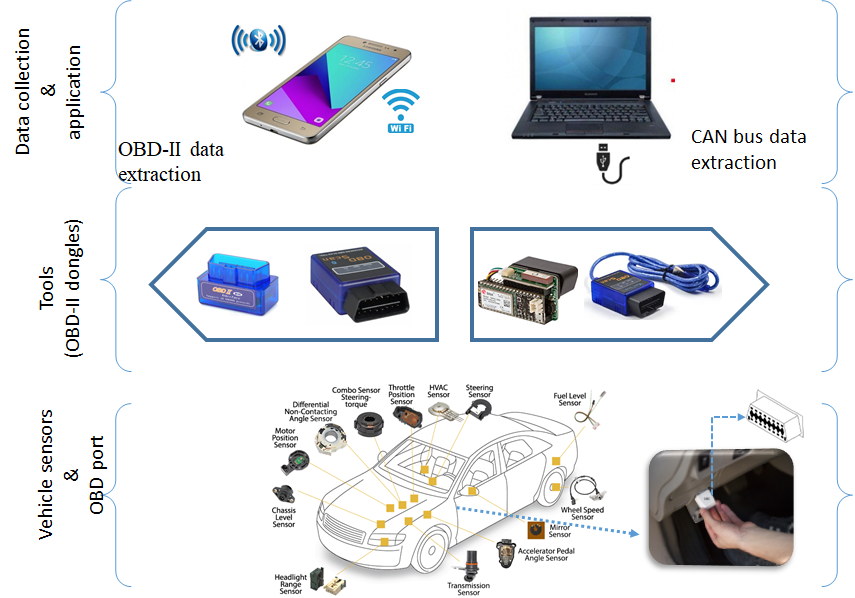}
	\caption{In-Vehicle (telematics) data acquisition through OBD-II interface}
	\label{fig:OBD}
\end{figure}

To address those mentioned vehicle security problems, this paper proposes a data-driven driver identification technique that can be implemented as an additional line of security for keeping cars safe from unauthorized drivers including thieves and hackers. The proposed method is based on freely available vehicle telematics data, also called OBD-II (On Board Diagnosis) data. OBD-II interface of a vehicle provides in-vehicle sensor reading such as vehicle speed, engine RPM, throttle position, engine load, break-pedal displacement, etc. As shown in Figure \ref{fig:OBD}, OBD-II dongles can be used to extract these internal sensors data in real-time to infer information about the car and its driver. In-vehicle sensors data are directly or indirectly are influenced by the drivers driving style. The driving style of each individual varies depending on how they maneuver their vehicle. How frequently the driver uses the brake and the gas pedal or how much pressure is applied on the brakes or how the steering wheel angle is adjusted at curves \cite{hallac2016driver} are a few examples. The driver's unique driving style attributes directly or indirectly manifested on generated vehicle telematics data. 

Combining vehicle telematics data with appropriate machine learning tools, helps one to recognize different driving styles \cite{hallac2016driver}, uncover driving behavior and patterns \cite{sathyanarayana2012leveraging}, and even detect hazardous driving behaviors \cite{imkamon2008detection}. However, most of the traditional algorithms used in the driver identification task rely on rigorous data-prepossessing steps that require either domain expert knowledge or an extensive data exploration process. Secondly, while the vehicle telematics sensor data is time-series data, conventional machine learning algorithms don't have the inherent ability to exploit sequential relationships. In contrast, end-to-end deep learning-based methods such as Recurrent Neural Network (RNN) can directly extract the most important features without data pre-processing and also exploit temporal relationship from the data in a holistic data-driven approach\cite{goodfellow2016deep,lecun2015deep}.


This paper proposes an end-to-end deep learning-based model as a driver identification technique using RNN architectures variant called Long-Short Term Memory (LSTMs). The end-to-end approach of LSTM enables to extract important features without rigorous data pre-processing procedures. The time-series nature of LSTM algorithm allows to holistically exploit the inherent temporal information embedded in time-series data captured from sensors in the driving sessions. Additionally, based on conducted studies, the proposed approach showed robust performance compared with other conventional machine learning algorithms even under the increasing influence of anomalous and noisy sensor data reading. Finally, we have made our model code and its' comparison with other models available at \href{https://github.com/Abeni18/Deep-LSTM-for-Driver-Identification-}{https://github.com/Abeni18/Deep-LSTM-for-Driver-Identification-}
The  main  contributions  of  this  paper  are  summarized  as follows:

\begin{itemize}
    \item We proposed a data-driven robust driver identification system based on end-to-end Long-Short-Term-Memory (LSTM)-Recurrent Neural Network model. The proposed model architecture utilizes a holistic data-driven approach to capture the driving signature of individuals out of telematics data to be able to identify the driver. 
    \item An efficient LSTM architecture is searched and implemented to achieve robust performance.
    \item The effect of sensor data anomalies and random environmental noise influence on the performance of the proposed approach is studied. We then compared the accuracy of the proposed driver identification model with three well-known conventional machine learning models, and a comprehensive comparison is presented. 
\end{itemize}

\indent The remaining part of this paper is arranged as follows: Section II reviews related work in the literature. Section III presents a detailed discussion of our proposed methodology. Section IV discusses the experimental studies on a real-world data-set, and the results are presented in Section V. Finally, the conclusions including summary of major points done in Section VI.

\section{Related Works}
With the success of machine learning algorithms and data mining techniques, in the last couple of years, a growing interest is shown in utilizing OBD-II data for developing a driver identification system. This section provides a brief review of the work conducted by the researchers to extract vital information from this data to identify drivers. 

Virtual simulators have been used by various researchers to generate data that is similar to real vehicle internal sensor data. Wakita et. \cite{wakita2006driver} and Zhang et. \cite{zhang2014study} have used simulation in controlled routes and settings to collect data to develop driver identification predictive models. Zhang et. used a hidden Markov model (HMM) to model individual characteristics of driving behavior based on accelerator and steering wheel angle data and managed to reach a maximum prediction accuracy of 85$\%$. On the other hand, Wakita et. used a Gaussian Mixture Model (GMM) with input data of accelerator pedal, brake pedal, vehicle velocity, and distance from a front vehicle and achieved 81$\%$ accuracy with twelve drivers driving in a simulator and 73$\%$ accuracy with $30$ drivers driving in an actual car. Both of these studies rely on few numbers of features or sensor readings, that could not be able to capture all the hidden behavioral information. Although these works provide insights into the success of machine learning approaches for this task, the results are not satisfactory enough or cannot be compared to the real-world situation that involves various uncontrolled settings such as different traffic patterns and environmental conditions like weather.

Kwak \cite{kwak2016know} studied the driving patterns of drivers using data collected in an uncontrolled environment from three road types; motorway, city way and parking lot with ten drivers participating in the experiment. This study compared Decision Tree, KNN, Random Forest and Fully Connected Neural Network algorithms based on their performance in predicting the drivers accurately. They performed data preprocessing that includes feature selection and statistical feature formation such as mean, median, and standard deviation to increase the model performance. According to the results, Random Forest and Decision Tree are the two most accurate algorithms\cite{kwak2016know, martinelli2018human}. They also showed the importance of adding statistical features to the data to get an accuracy level above 95$\%$. 

Fabio \cite{martinelli2018human} classified drivers through a detailed analysis of human behavior characterization for driving style recognition in-vehicle systems. The pre-processed data is used to compare five classification algorithms J48, J46graft, J47consolidated, Random-tree, and Rep-tree. According to their results, J48 and J48graft classification algorithms showed better performance on different measurement matrices. This work focused on the statistical analysis of the data by comparing different classification algorithms based on various performance metrics.

\begin{figure*}
    \centering
    \includegraphics[width=6.6in,height=1.8in]{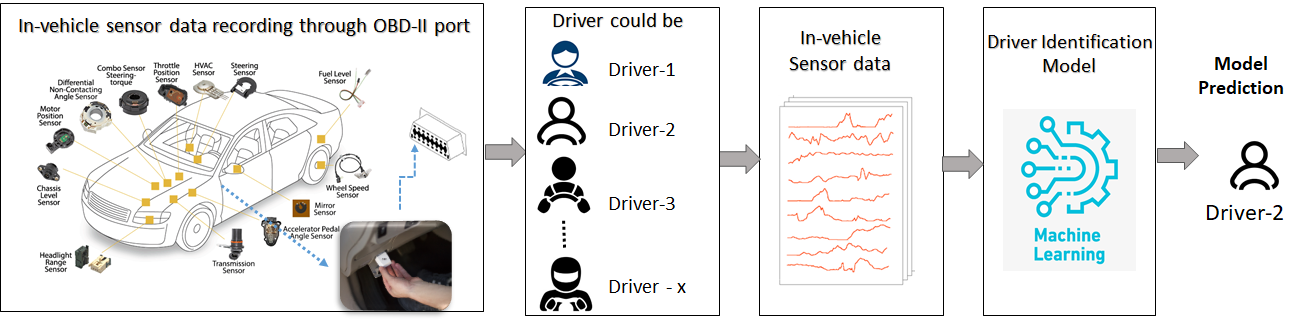}
    \caption{Driver Identification System Framework}
    \label{fig:system}
\end{figure*}

		
Most of these studies in literature focused on rigorous data analysis that involves feature selection  \cite{yan2019novel,yan2018unsupervised}, and statistical feature formation techniques. These processes are time-consuming, and some features may only exist in some vehicles, which limits the practicability of the work. In this paper, we do not use feature selection and formation process. By taking raw data from any car, our approach can extract important features in a holistic data-driven approach. Secondly, while the OBD data has a time-series property, they haven't used time-series algorithms. This hinders the advantage of using embedded temporal information in the data. Thirdly, even though the effect of anomalous and noisy sensor reading is a common problem in real-world implementation, their effect on the performance of the algorithms used in driver identification system hasn't been studied.

\section{ Proposed Methodology}
This section discusses how we developed a driver identification model based on OBD data to provide an additional line of security for vehicles' protection. Firstly, the problem formulation of the driver identification system is discussed. Secondly, an explanation of the model and the learning process we adopted to solve the problem are presented. Finally, more details about the proposed method is discussed. 

\subsection{Problem Formulation}
\begin{figure}
    \centering
    \includegraphics[width=3.5in,height=2.2in]{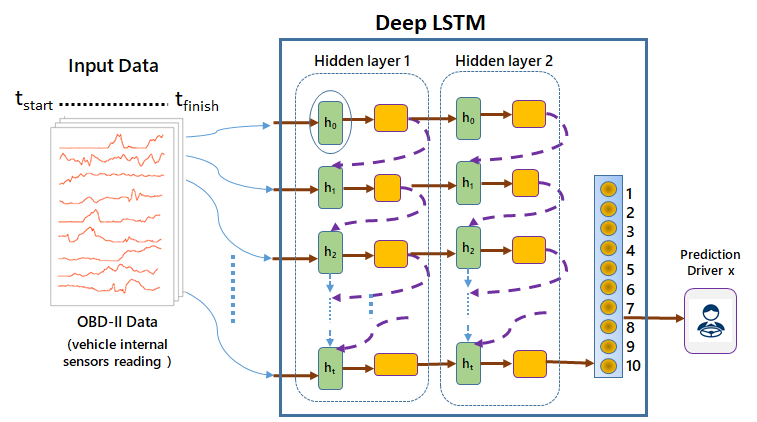}
    \caption{Deep LSTM Driver Identification model Architecture}
    \label{fig:arch}
\end{figure}

Given OBD-II data is a sequence of sensor data collected over time from the car during a driving trip, we formulate the driving identification problem as a time-series prediction task where a window sequence of OBD-II data $'S'$ from time $\boldsymbol T_{start}$ to time $\boldsymbol T_{finish}$ can be identified as it is from one of the drivers out of a given set of individuals, as shown in Figure \ref{fig:system}, \ref{fig:arch}. Additionally, the performance of the model to give a correct prediction under different level of environmental noise influence and sensor anomalies has been investigated.  

\subsection{Model development}
\label{Section III-B}
Recurrent Neural Networks (RNNs) are one of the most successful time series algorithm techniques that have been widely used for time-series classification tasks like speech recognition, machine translation, and human activity recognition \cite{cho2014learning}. As shown in Fig. \ref{fig:arch}, RNNs take sequential input $\boldsymbol X_{t}$ at time $\boldsymbol t$ and output $\boldsymbol y_{t}$ based on a decision made by an internal state $\boldsymbol h_{t}$. The internal state is a connection used to extract important features and capture the temporal dynamics of time-series data,i.e., data coming from sensor reading. This internal state, named \textbf{hidden state}, is multiple copies of a fully connected neural network, each passing a message to a successor for the next time step. 

Based on the number of inputs and expected outputs, RNNs architectures can be designed in different ways. In this work, there are multiple input features at the input layer, and a single prediction output is expected at the output layer. Because of that \textbf{``Many to One"} type of RNN architecture is used called \textbf{``Many to One"}.  As shown in Figure \ref{fig:arch}, our model takes  $\boldsymbol X_{s}$  feature vectors of time-series data with a window size of ($\boldsymbol  X^{0}$ to $\boldsymbol  X^{n}$ ) and predicts output vector $\boldsymbol O_{s}$. Then out of the predicted output, the one with the highest probability is selected as a final prediction.
Out of available features, at each time step, one feature vector is analyzed by the model, which makes the number of time steps equal to the number of features in the data. While the data passes through the hidden layer of Deep-LSTM neural network, important features are extracted, and temporal dependencies get captured automatically. Finally, after the last time-step, the network produces a likelihood score for each driver as an output value. Then, the one with the highest score is selected as a prediction. Unfortunately, the ``vanishing gradient" problem can cause RNNs to become unable to learn to connect long term dependencies when the number of the hidden state increases.

One particular variant of RNNs architectures called the \textbf{Long-Short Term Memory (LSTM)} network addresses this problem by using "memory block" in the hidden unit to capture the long-term-dependencies that could exist in the data. This memorizing capability of LSTM has shown the best performance in many time-series tasks such as activity recognition, video captioning, language translation \cite{ordonez2016deep,sundermeyer2012lstm}.

The cell state (memory block) of LSTM has one or more memory cells that are regulated by structures called gates. Gates control the addition of new sequential information and the removal of useless to and from memory, respectively. Gates are a combination of sigmoid activation function and a dot (scalar) multiplication operation, and they are used to control information that passes through the network. An LSTM often has three gates, namely forget, input, and output gates, which are summarized as follow:

\begin{figure}[!h]
	\centering
	\includegraphics[width=2.8in,height=2.0in]{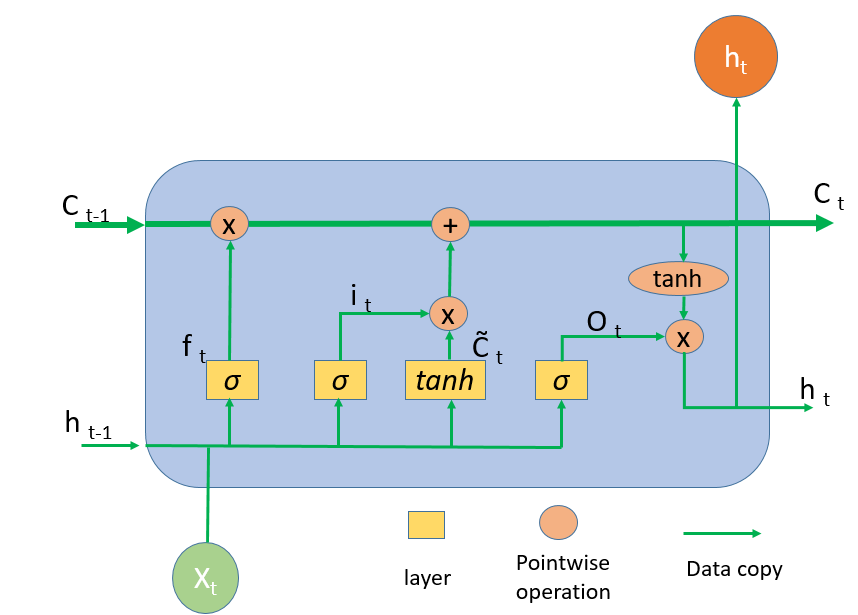}
	\caption{Long-Short Term Memory (LSTM) memory block graphical representation.}
	\label{fig:2}
\end{figure}
\begin{itemize}
	\item \textbf{Forget gate} : Forget gate, equation \ref{eq:forget_gate}, decides what information to keep or remove from the cell state. 
	\begin{equation}
	f_{t} = \sigma(W_{f}.[h_{t-1}, x_{t}] + b_{f})  
	\label{eq:forget_gate}
	\end{equation}
	\item \textbf{Input gate} : Input gate, equation \ref{eq:input_gate}, decides what new information to add and how  to update the old cell state, $C_{t-1}$, to the new cell state $C_{t}$ for the next memory block. 
	\begin{equation}
	\begin{split}
    i_{t} = \sigma(W_{i}.[h_{t-1}, x_{t}] + b_{i}) \\
    \hat{C}_{t} = tanh(W_{c} . [h_{t-1}, x_{t}] + b_{c}  ] \\
    C_{t} = f_{t} * C_{t-1} + i_{t} *\hat{C}_{t}
    \end{split}
    \label{eq:input_gate}
    \end{equation}
	\item \textbf{Output gate} : Finally output gate, equation \ref{eq:output_gate}, filters out and decides which information to produce as an output from a memory block at a given time step t.
	\begin{equation}
	\begin{split}
	o_{t} = \sigma(W_{o}[h_{t-1}, x_{t}] + b_{o}) \\
	h_{t} = o_{t} + tanh(C_{t})
	\end{split}
	\label{eq:output_gate}
	\end{equation}

\end{itemize}

\begin{table}[!h]

\centering
\caption{Table of Parameters for the LSTM model}
\begin{tabular}{|l|l|}
\hline
 Variables & Definition (respectively)    \\ \hline
$X_{t}$ and $h_{t}$  &  input and output of the memory cell\\ \hline
$h_{t-1}$  & input from previous state \\ \hline
$f_{t}$, $i_{t}$, $o_{t}$  & activation function of forget,input $\&$ output gates  \\ \hline
$W_{f}$, $W_{i}$,  $W_{C}$, $W_{o}$, & weights of forget,input, candidate $\&$ output gates  \\ \hline
$b_{f}$, $b_{i}$, $b_{c}$, $b_{o}$ & biases of forget, input, candidate and output gates\\ \hline
${\hat{C}}_{t}$ and ${C}_{t}$   & candidate cell and updated cell state value \\ \hline
\end{tabular}

\end{table}

The memory blocks are connected to build layers of the Deep LSTM neural network, as shown in Figure \ref{fig:arch}. The last layer of our Deep-LSTM model is a sigmoid function which is shown in equation \ref{eq:sigmoid}. It takes the last hidden layer feature vectors and outputs classification scores for the given set of drivers. The one with the highest probability score is then selected as a final prediction. During training, a family of the cross-entropy loss function and an Adam optimization algorithm are used. These are some of the hyper-parameters that are fixed initially to build the model. There are other hyper-parameters such as the number of neurons, the depth of the network, the input data sequence length, and the data overlap amount. The best combination of these hyper-parameters has to be selected for developing an efficient model. Accordingly, several experiments are conducted to choose those hyper-parameters, which is discussed in the next section. 

\begin{equation}
\sigma(z) =  \frac{1} {1 + e^{-z}} 
\label{eq:sigmoid}
\end{equation}
Where z is the output of the network.

\subsection{OBD-II Data}
European and North American countries adopted OBD (On Board Diagnosis) technology to standardize the way vehicles can be checked for compliance. The OBD-I system was firstly used in the early 1980s, and since 1998 all vehicles sold in the USA were enforced by law to be OBD-II prepared \cite{obdmanual}. There are several standardized communication protocols used by car manufacturers to communicate data between ECM (Electronic Control Modules) of a car and a scan tool such as ISO 9141-2, SAE J1850 PMW, SAE J1850 VPW, ISO 14230-4, SAE J2284, ISO 15765-4, etc \cite{obdmanual}. OBD-II scan tool can automatically detect the communication protocol and vehicle features which makes it a plug-and-play device.

In comply with OBD-II standards, OBD program is divided into sub-group programs referred to as 'Service \$xx', which range from Service \$01-to-Service\$09 and Service \$0A. Accordingly, service \$01 displays an actual real-time reading of in-vehicle sensor data, where the data is commonly referred to as PID (Parameter Identification) data. Other services such as Service \$02 displays the state of the PIDs when a fault occurred, Service \$03 used to access DTCs (Diagnostic Trouble Code) to display trouble codes and so on. Service \$01 is an actual sensor reading where most of the features commonly exist across different cars, and it's also real-time information. As a result, our approach mainly used data obtained from Service \$01 as a source of information.


\section{Experiments}
The model development and experimentation process consist of three steps, which are data preparation, LSTM architecture design, and model robustness study on dataset affected by sensor data anomalies and noise. 

\subsection{Data Description and preparation}
As shown in Figure \ref{fig:OBD} and \ref{fig:system}, Bluetooth, WiFi or USB enabled OBD-II dongles can be plugged into the  OBD-II interface of the car to collect the internal real-time sensors reading by external devices like mobile phone, laptop or cloud. The experiments conducted in this paper used the following three datasets:

\indent \textit{Security Driving Dataset}\cite{security_dataset}: This dataset is collected by KIA Motors Corporation car with CarbigsP as OBD-II scanner. The experiment took place in an uncontrolled environment that consists of three types of path; city way, parking space, and motorway. Ten drivers participated in the experiments, and for a reliable classification, each driver completed two round trips on weekdays during off-peak hours from 8 p.m to 11 p.m. There are $94,401$ data points recorded with $51$ different independent features.

\indent \textit{Vehicular data trace Dataset-1} \cite{vehicular_tarce}: This dataset is collected as a part of Intelligent Transportation Systems (ITS) study and Vehicular Ad-hoc Networks (VANETs). The data is obtained from the OBD-II interface via Bluetooth connection to an Android app installed in a smart-phone. In this data collection, the Hyundai HB20 model vehicle is shared by ten drivers for 36 trips covering a total trip time of 28 hours in their daily routines. Therefore, the experiment is naturalistic and uncontrolled. Six male and 4 females with an age range of 25 to 61 participated in the experiment.

\indent \textit{Vehicular data trace Dataset-2} \cite{vehicular_tarce}: This dataset is collected in a similar situation as Vehicular data-trace Dataset-1 except the experiment is controlled. In this experiment, a Renault Sandero model vehicle is shared with four drivers to drive through two different selected routes for a total trip time of 3 hours, which makes it a controlled experiment. Two male and two female with an age range of 20 to 53 participated in the experiment.

To cancel out the effect of different sensor measurement scales, the datasets are standardized using equation \ref{eq:norm}, no further preprocessing is applied. Then the datasets are split into training 85$\%$, validation 5$\%$ and test 10$\%$ data. All training, validation, and test data are then sliced into sequences of a sliding windows data chunk. The study used  50$\%$ overlapping between a consecutive series of data chunks before the data is fed to the model. The overlapping window is important to smooth the flow of sequence, capture sequential information, and increase training data size for better generalization. 
\begin{equation}
X_{new} = \dfrac{X - X_{min}}{X_{max} - x_{min}}
\label{eq:norm}
\end{equation}
Where X denotes a given data point, $X_{min}$ and $X_{max}$ are minimum and maximum data points in the data set and $X_{new}$ is the new normalized data. 
\begin{figure}[!h]
	\centering
	\includegraphics[width=3.5in,height=1.3in]{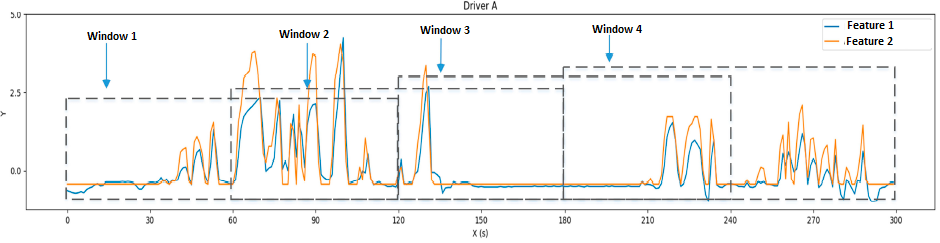}
	\caption{Overlap sliding window method between consecutive sequence of data of feature 1 and 2}
	\label{fig:seq}
\end{figure}

\subsection{Deep-LSTM architecture design}
As discussed in the previous section, `many-to-one' Deep-LSTM formation is appropriate for this problem. As shown in Figure \ref{fig:arch}, `Many-to-One" only defines the number of input(many) and output(one) of the network, whereas the internal part of the network could be constructed in several ways. The number of layers and the number of neurons in hidden layers are some of the internal network parameters that determine both the accuracy and computational complexity of the model \cite{goodfellow2016deep}. As the depth of the network and number of neurons in the hidden layer increases the accuracy often increases, but at a cost of computational resources \cite{lecun2015deep}. A grid architecture search technique is applied using training and validation data to find the most efficient architecture. A two hidden layer network with  $160$ neurons in the first hidden layer and $200$ neurons in the second hidden layer found to be the efficient neural architecture. 

\begin{figure}[!h]
	\centering
	\includegraphics[width=2.5in, height=1.5in]{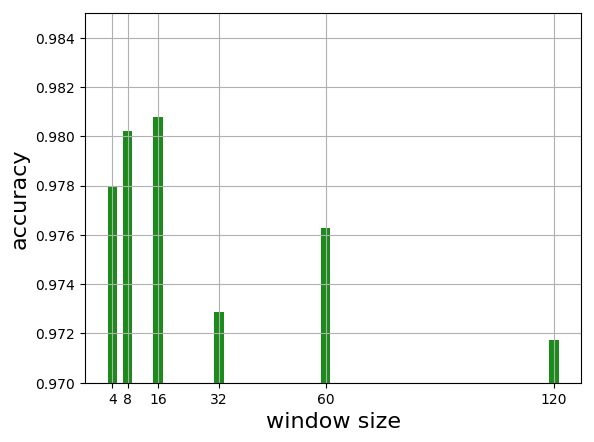}
	\caption{Search for window size for sequence of data}
	\label{fig:window}
\end{figure}

As a time series algorithm, LSTM model takes a sequence of data as an input. Accordingly, we searched for an adequate number of sequence size. With 85$\%$ training and 5$\%$ validation data, different window sizes (4, 8, 16, 32, 64 and 120 data points) are tested on the developed architecture. As shown in Fig. \ref{fig:window}, a window size of 16 data-points is found to be the best fit. In the next section, the performance of the designed architecture is tested on real-world data-set, and its robustness against noise and sensor anomalies is studied.
\subsection{Model Robustness to sensor data anomalies and noise } \label{Section IV-C}
The performance of telematics data-based driver identification systems depends on the reliability of the collected vehicle sensors data. However, sensors are prone to failures, defects, and cyberattack attempts which introduces anomaly in the data \cite{narayanan2016obd_securealert, ni2009sensor, hill2007real}. Additionally, due to extreme environmental factors such as temperature, noise could also significantly and constantly corrupt the quality of sensor data during generation and transport the data to the central data collection unit \cite{kalapanidas2003machine}. Thus, the driver identification model has to be robust in such anomalous and noisy data to ensure a correct prediction in a real-world implementation. By considering this, we have studied the effect of increasing levels of sensor data anomalies and noise influence on the performance of the proposed approach and its comparison with other three widely used machine learning algorithms. 

\begin{figure}[!h]
    \subfigure{
    \includegraphics[width=1.5in, height=1.5in]{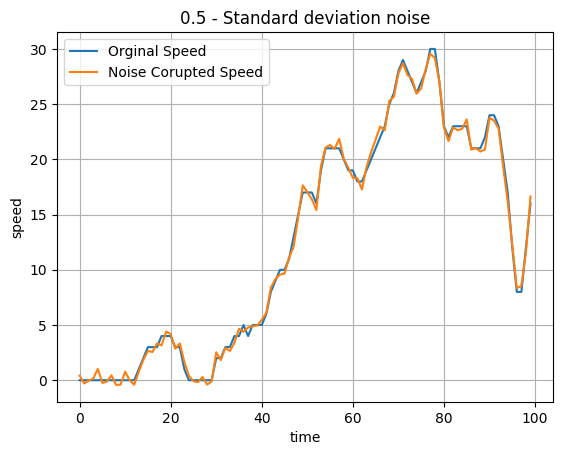}
	}
	\subfigure{
	\includegraphics[width=1.5in, height=1.5in]{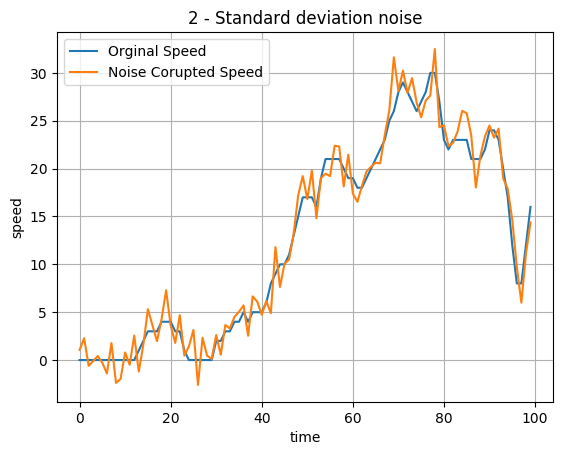}
	}
	\caption{Visualization of dataset inflicted with different level of noise }
	\label{fig:noise_plot}
\end{figure}

\begin{figure}
    \centering
    \subfigure{
    \includegraphics[width=1.5in, height=1.3in]{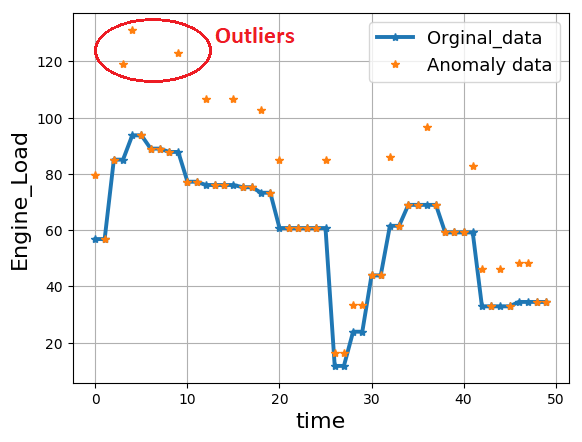}
	}
	\subfigure{
	\includegraphics[width=1.5in, height=1.3in]{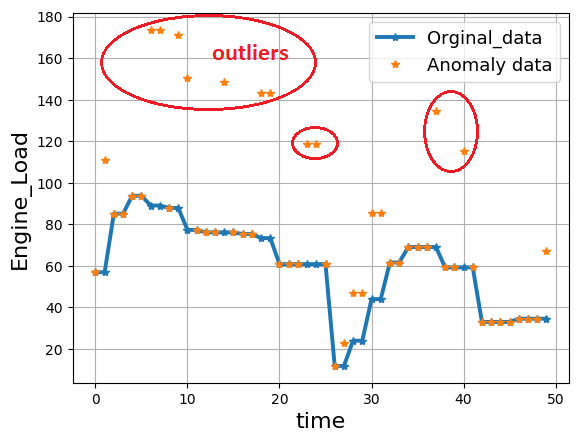}
	}
    \caption{Increasing level of anomalies rate induced in the data}
    \label{fig:anomal}
\end{figure}

Noise is a random error or variance in a measured variable \cite{han2000m}. In this study, White Gaussian Noise (WGN) type is applied to degrade the original dataset at hand. WGN is used to mimic the effect of random noise occurrence in electronic system \cite{kalapanidas2003machine, han2000m}. It is assumed that the noise is randomly distributed for all models, and it is independent of the original data. Then every independent variable $X_{old_i}$ in dataset is going to be substituted by a noise inflected one $X_{new_i}$ with a probability of n, where n refers to the level of noise. Equation \ref{eq:noise_eq} is used to calculate the new noisy data. 
\begin{equation}
X_{new_i} =\begin{cases} 
    X_{old_i} + rand( \sigma_{i} , \mu_{i} ) & p_{i} >= n , \\
    X_{old_i}  &  p_{i} < n
   \end{cases}   
\label{eq:noise_eq}
\end{equation}
Where the noise is generated with probability $p_{i}$ using a random signal (rand) that has zero mean $\mu_{i}$, which makes it centered on the original data value, and standard-deviation based on the data variance $\sigma_{i} $ that describes how severe the noise affects the data. The induced standard deviation (std) varies from zero std (no noise) to two std, we choose two in order to keep the noise effect in two normal distribution region. When the noise is added on the data it keeps the original patterns but it wiggles individual data points away to some extent from the actual value based on the standard deviation of the noise. As shown in figure \ref{fig:noise_plot} the harshness level of the noise can be increased by increasing the standard deviation of the noise.
On the other hand anomalies/outliers are abnormal extreme patterns or events in data that do not conform to a well defined notion of normal behavior \cite{hayes2015contextual}. Most observations in data lies in between two normal regions (standard deviation), points that are far away from these regions are usually referred to as anomalies or outliers \cite{chandola2009anomaly}. Accordingly, we induced random anomaly in the original data that push some of the data points out of the normal region by increasing the values of data points with the given percentage. As shown in Figure \ref{fig:anomal}, when 40 \% of Engine load sensor reading data is affected by 40 \% of anomaly (40\% increment from original data point) and 85 \% of anomaly, some of the affected data point changed to outliers. In the next sections, the performance of the proposed approach will be examined and tested under the influence of such induced noise and sensor anomalies. 

\section{Results and Discussions}
Three well-known evaluation measures, namely F1-Score, Precision, and Recall, are used to evaluate the performance of the proposed Deep LSTM model. These metrics are defined in the following equations. 
\begin{itemize}
    \item Precision:
    \begin{equation}
    Precision = \dfrac{TP}{TP + FP}.
    \end{equation}
    \item Recall:
    \begin{equation}
    Recall =  \dfrac{TP}{TP + FN}.
    \end{equation}
    \item F1-score:
    \begin{equation}
    F1  = 2 \times \dfrac{Precision \times Recall}{Precision + Recall}.
    \end{equation}
\end{itemize}
Where $TP$ denotes the number of samples that has the same predicted label with the true class label, $FP$ represents the number of samples that are classified into a class that does not belong to original class. The term $FN$ refers to the number of samples that the classifier fails to classify. In this study, TensorFlow \footnote{https://www.tensorflow.org/} \& Keras\footnote{https://keras.io/} Deep Learning Library is used to develop our LSTM model, and sklearn \footnote{https://scikit-learn.org/} machine learning library is used to replicate other algorithms used for compression. The computer used for developing, training and testing the models is Hp Z840 workstation with Intel Xeon CPU and 64 GB RAM size.

Three datasets are used to evaluate the proposed approach, where the datasets are separately divided into 85\% training and 5 \% validation and 10 \% test data. From TABLE \ref{table:accuracy}, it is observed that the proposed LSTM model provides high accuracy in predicting driver identity. For instance, the average recall, precision, and the F1-Score of the proposed LSTM model are above $97\%$, which indicates the efficacy of the model in the driver identification tasks.

\begin{table}[!h]
  \caption{LSTM model accuracy on naturalistic driving datasets}
	\begin{tabular}{|l|l|l|l|l|}
		\hline
		
		Dataset &  Drivers  & Precision & Recall  &F1 score  \\ \hline
		Security Driving data & 10         &  0.988  &  .981 & .98   \\ \hline
		Vehicular  data  trace -1&    10       &    0.97 &  .972 & .975    \\ \hline
		Vehicular  data  trace-2&   4       &   0.99   &   0.991 & .987   \\ \hline
	\end{tabular}
  
    \label{table:accuracy}
\end{table}

	
		


\begin{figure}[!h]
	\centering
	\includegraphics[width=3.1in, height=1.8in]{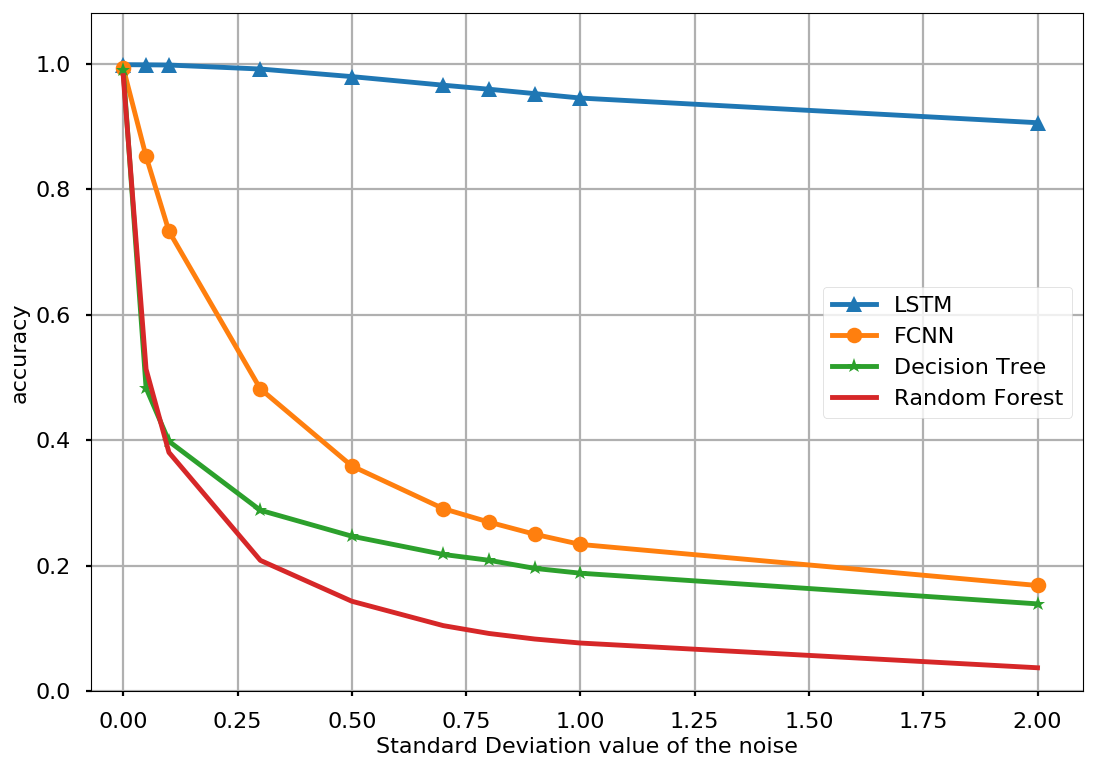}
	\caption{Models accuracy comparison by inducing increasing level of noise on test-data from Security Driving Dataset}
	\label{fig:acc_noise}
\end{figure}

To compare our model against other driver identification techniques, we picked the three most popular machine learning algorithms from literature, which are Random Forest (RF), Decision Tree (DT) and Fully Connected Neural Network (FCNN). The same dataset is used by other authors is also used in our study \cite{kwak2016know,zhang2016driver, martinelli2018human}. The same data split is used to train and test all models that are training 85\%, validation 5\%, and test 10\% of the overall data. As discussed in section \ref{Section IV-C}, in the first experimental study,  all models are trained on training data with no noise, and then they are evaluated on test data with noise. White Gaussian Noise (WGN) is applied to the dataset where standard deviations of the WGN are used to control the induced level of noise.
Each model is separately evaluated on test data with different noise level varying from zero levels to two standard deviation level (in normal distribution range). The model is evaluated ten times for each noise level, then the average accuracy value is taken, and the result is presented in Fig. \ref{fig:acc_noise}.
 

\begin{figure}[!h]
	\centering
	\includegraphics[width=3.in, height=1.5in]{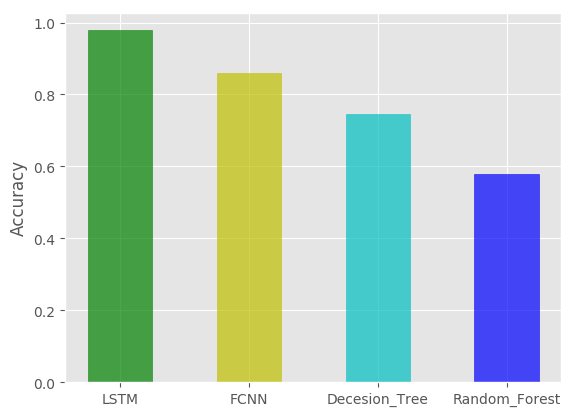}
	\caption{Models accuracy comparison trained and tested on noise inflicted Security Driving Dataset }
	\label{fig:acc_noise_train}
\end{figure}
\begin{figure}[!h]
	\centering
	\includegraphics[width=3.2in, height=2.7in]{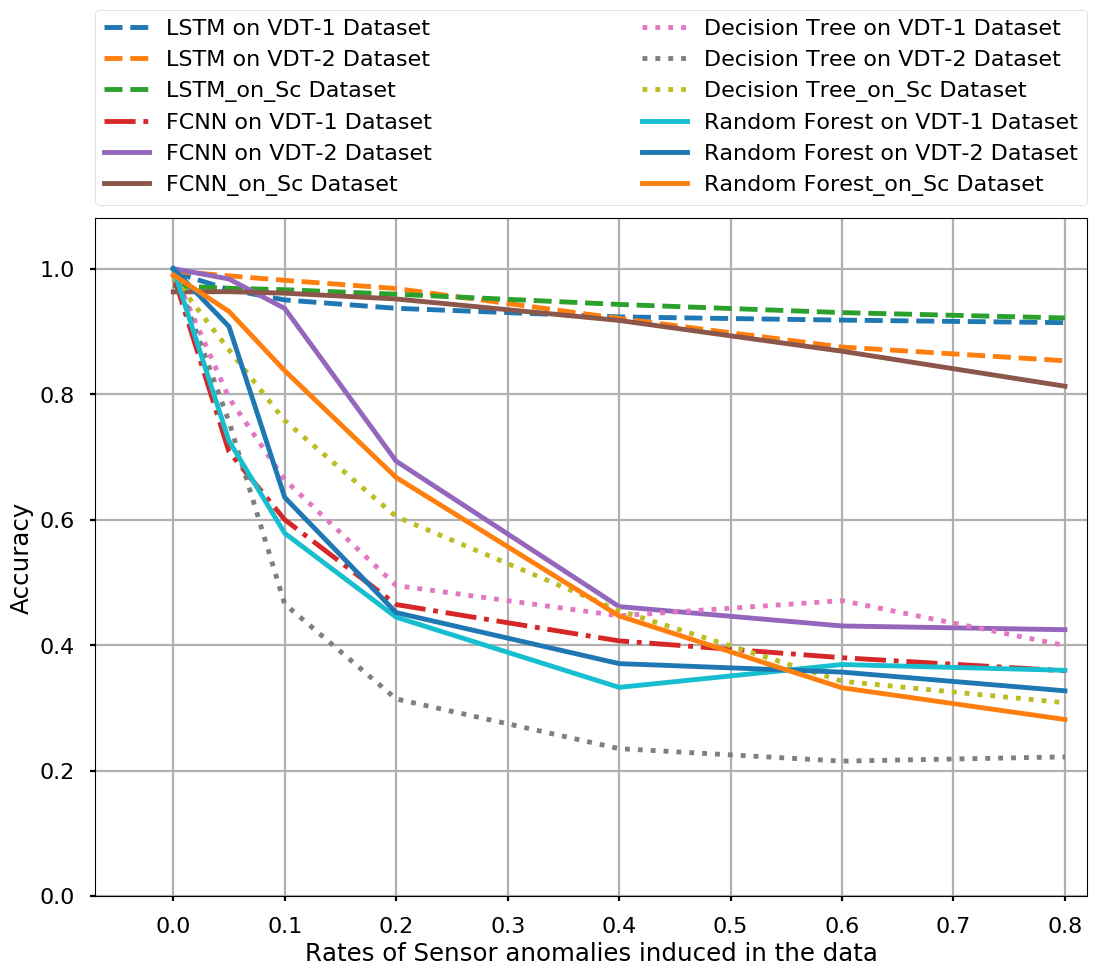}
	\caption{Models accuracy on an increasing level of anomalies induced on VDT(Vehicular data-trace) and Security (Sc) dataset}
	\label{fig:acc_anomal2}
\end{figure}

On the other hand, modeling a supervised machine learning technique that can effectively learn from dataset already inflicted with noise is a problem of great practical importance \cite{natarajan2013learning}. In practical applications, it is observed that models usually overfit in the presence of noise in training data \cite{manwani2013noise}. To train driver identification model in real-time on unclean data directly coming from the vehicle, the model needs to be robust enough to noises in training data too. Accordingly, in the second experimental study,  we further experimented by training the selected models on noisy data and then testing them on noisy data. As shown in Figure \ref{fig:acc_noise_train}, neural network-based models (LSTM followed by FCNN) performed better than the others by avoiding over-fitting on the noise.
\begin{figure}[!h]
	\centering
	\includegraphics[width=3.2in, height=2in]{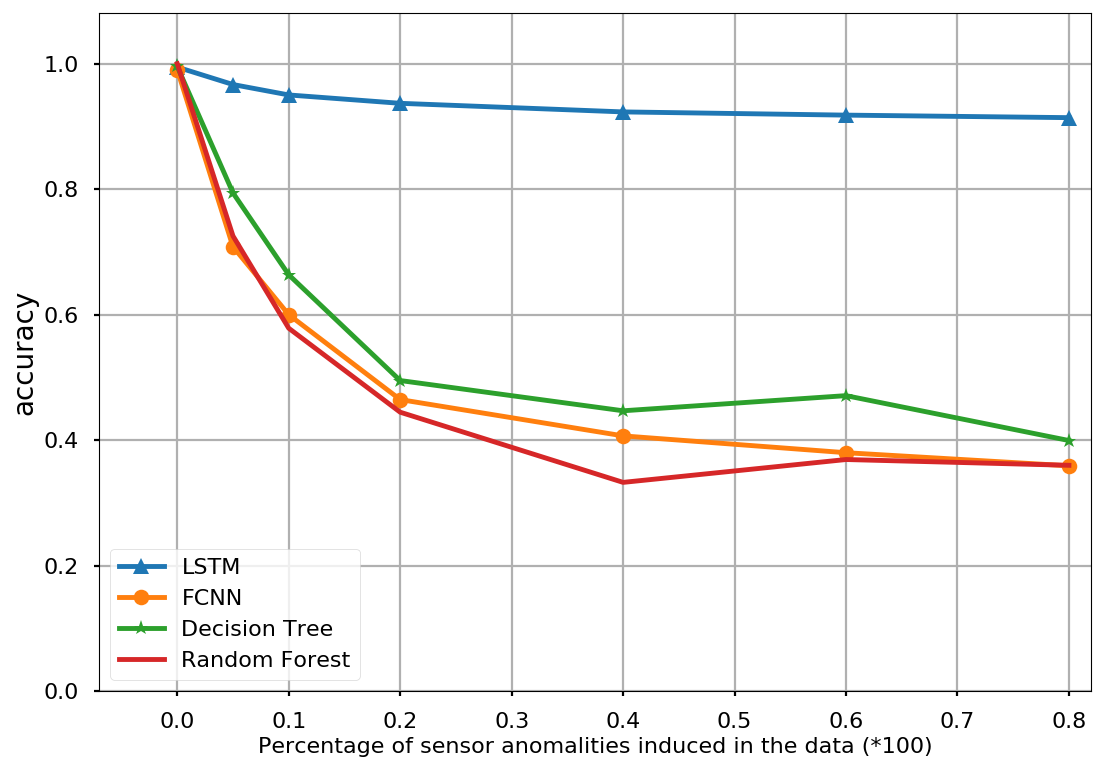}
	\caption{Models accuracy on increasing level of anomalies induced on test-data from Vehicular data trace Dataset-1}
	\label{fig:acc_anomal1}
\end{figure}
As discussed in Section \ref{Section IV-C}, aiming to assess the impact of anomalies/outliers of sensor data on the performance of the models, we introduced a different level of outliers to the original test-data. Accordingly, in the third experimental study, 40 \% of overall test-data affected by anomaly rate ranging from 0\% to 65 \%. As shown in Figure \ref{fig:acc_anomal1} and \ref{fig:acc_anomal2}, a comparison between the proposed approach accuracy obtained on anomalous data against other models is presented.

As presented in the above results, for clean data, conventional machine learning models used in literature's have comparable or less accuracy compared to the proposed model. However, under sensor anomalies/outliers or environmental noise, other model's performance quickly droped below an unacceptable range. But, the proposed Deep-LSTM model keeps its accuracy above an acceptable level in all cases. This difference is mainly attributed to the fact that unlike LSTM, other classical machine learning models do not have an inherent ability to exploit temporal relationship from time-series data. Other conventional machine learning models, including FCNN, examines a data point recorded at one-time step (single row). But, LSTM examines a sequence of data points (multiple consecutive rows) to extract time-dependent patterns from the data. The internal memory in LSTM helps to remember patterns existed in previous time steps in relation to current timestep or events.  As a result, LSTM  is considered to be the best machine learning algorithm at remembering pieces of information and keep it saved for many time steps \cite{lecun2015deep}. The importance of capturing the temporal relationship is clearly shown when we compare the two models from the same family, which are Fully Connected Neural Network (FCNN) and the proposed LSTM Recurrent Neural Network. Even though the same exact network was used for building both neural networks architecture, LSTM performed well under noise; however, FCNN performed poorly. Therefore, due to the above-discussed reasons and its architectural design (this discussed in detail in section \ref{Section III-B}) the proposed LSTM based approach performed better than the other conventional machine learning models.

\section{Conclusions}
To address an increasing vehicle security problem, we presented an end-to-end LSTM-RNN architecture as a driver identification model. The model is developed based on freely available vehicle telematics data collected from the OBD-II interface of vehicles.  The problem is formulated as a time-series prediction task, where the model is trained on a sequence of in-vehicle sensor data. LSTM has an inherent ability to remember temporal information in data and keep it saved for many time steps than the other conventional machine learning approaches.  Accordingly, the proposed model efficiently learns individual unique driving patterns from the data to identify the driver. The holistic data-driven approach of the technique also has the advantage of avoiding rigorous data pre-processing procedures. The proposed method is evaluated on a real-world dataset using different metrics and achieved a better or comparable result against other models. Further studies on anomalous and noisy sensor data show our model scores substantially better than the others. Even under increasing noise and outliers effect, the proposed approach maintains its accuracy above the acceptable value, 88\%, while other models' accuracy goes below 40$\%$.

\section{Acknowledgement}
This work is based on research supported by NASA Langley Research Center under agreement number C16-2B00-NCAT, and Air Force Research Laboratory and Office of the Secretary of Defense (OSD) under agreement number FA8750-15-2-0116.

\bibliographystyle{unsrt}
\bibliography{mybibfile}

\end{document}